\documentclass{article}

\usepackage[final]{IIBPSL_2020}
\usepackage[numbers]{natbib}

\usepackage[utf8]{inputenc} % allow utf-8 input
\usepackage[T1]{fontenc}    % use 8-bit T1 fonts
\usepackage{hyperref}       % hyperlinks
\usepackage{url}            % simple URL typesetting
\usepackage{booktabs}       % professional-quality tables
\usepackage{amsfonts}       % blackboard math symbols
\usepackage{nicefrac}       % compact symbols for 1/2, etc.
\usepackage{microtype}      % microtypography

\usepackage{arydshln}
\usepackage{placeins}
\usepackage{amsmath}
\usepackage{cleveref}
\usepackage{amssymb}
\usepackage{breqn}
\usepackage{float}
\usepackage{printlen}
\usepackage{wrapfig}
\usepackage[bottom]{footmisc}
\usepackage{enumitem}
\usepackage{soul}
\usepackage{caption}
\usepackage{subcaption}
\usepackage{arydshln}
\usepackage{booktabs}
\usepackage{tikz}
\usepackage[most]{tcolorbox}
\usepackage[export]{adjustbox}
\usepackage{float}
\usepackage{hyperref}
\usepackage{url}
\usepackage{appendix}

\usepackage{amsmath,amsfonts,bm}

\def\eqref#1{equation~\ref{#1}}
\def\1{\bm{1}}

\DeclareMathAlphabet{\mathsfit}{\encodingdefault}{\sfdefault}{m}{sl}
\SetMathAlphabet{\mathsfit}{bold}{\encodingdefault}{\sfdefault}{bx}{n}

\newcommand{\R}{\mathbb{R}}

\title{Simulating Surface Wave Dynamics with Convolutional Networks}

\author{%
  Mario Lino \\
  Department of Aeronautics\\
  Imperial College London\\
  \texttt{mal1218@ic.ac.uk} \\
  \And
  Chris Cantwell \\
  Department of Aeronautics \\
  Imperial College London \\
  \And
  Stathi Fotiadis \\
  Department of Bioengineering  \\
  Imperial College London \\
  \And
  Eduardo Pignatelli \\
  Department of Bioengineering  \\
  Imperial College London \\
  \And
  Anil Anthony Bharath \\
  Department of Bioengineering  \\
  Imperial College London \\
}

\begin{document}

\maketitle

\begin{abstract}
We investigate the performance of fully convolutional networks to simulate the motion and interaction of surface waves in open and closed complex geometries. We focus on a U-Net architecture and analyse how well it generalises to geometric configurations not seen during training. 
We demonstrate that a modified U-Net architecture is capable of accurately predicting the height distribution of waves on a liquid surface within curved and multi-faceted open and closed geometries, when only simple box and right-angled corner geometries were seen during training.
We also consider a separate and independent 3D CNN for performing time-interpolation on the predictions produced by our U-Net.
This allows generating simulations with a smaller time-step size than the one the U-Net has been trained for.
\end{abstract}

\section{Introduction}

\begin{wrapfigure}{R}{0.4\textwidth}
  \begin{center}
    \includegraphics[clip,width=0.4\columnwidth,trim={300mm 250mm 250mm 250mm}]{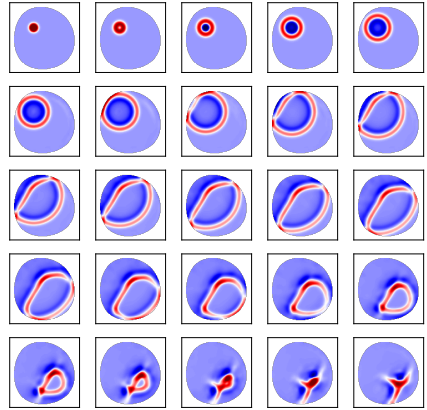}
  \end{center}
\caption{Rollouts of our U-Net. It simulates wave motion on a fluid surface with the possible existence of solid walls [\href{https://ibb.co/Cw4m4wt}{\underline{video}}].}
\label{fig:intro}
\end{wrapfigure}

We study the application of fully Convolutional Neural Networks (CNNs) to the problem of forecasting surface wave dynamics, the motion of which is described by the shallow water equations \citep{ersoy2017saint}. Computational modelling of surface waves is widely used in seismology, computer animation and flood modelling \citep{ersoy2017saint, garcia2019shallow}.
Our network learnt to simulate a range of physical phenomena including wave propagation, reflection, interference and diffraction at sharp corners.
This kind of Neural Network (NN) could supplement or potentially replace numerical algorithms used to solve the shallow water Partial Differential Equations (PDEs), reducing the inference time by several orders of magnitude and allowing for real-time solutions. This has particular relevance in iterative design scenarios and applications such as tsunami prediction.

Our NN produces the physics simulation by performing one time-step each time it is evaluated.
The time-step size of these simulations is fixed and must be equal to the one used during training; hence it is not possible to obtain predictions between time points.
This inconvenience is common to the vast majority of CNNs used for simulating physics \citep{lee2019data,kim2019deep,sorteberg2018approximating}.
To overcome this issue, we could train a NN to simulate within a range of time-steps sizes, but, this would imply extensive data sets and training time.
As an alternative, we suggest the use of two independent models: a NN for predicting the temporal evolution with a fixed time-step size, and a second NN for performing the time-interpolation on such simulation.

\textbf{Contribution.}
We demonstrate that our U-Net architecture is able to accurately predict surface wave dynamics in complex straight-sided and curved geometries, even when trained only on data sets with simple straight-sided boundaries.
Our U-Net is able to simulate wave dynamics four orders of magnitude faster than a state-of-the-art spectral/$hp$ element numerical solver \citep{spencer}, so it could be an effective replacement for numerical solvers in applications where performance is critical.
We also demonstrate that a 3D CNN is able to time-interpolate our U-Net predictions and increase the temporal resolution of the simulations by a factor of four.

\section{Related Work}
During the last five years, most of the networks used to predict continuous physics have included convolution layers.
For instance, CNNs have been used to solve the Poisson's equation \citep{Tang2018,girayhan2019poisson}, and to solve the steady Navier-Stokes equations \citep{guo2016convolutional,miyanawala2017efficient,yilmaz2017convolutional,farimani2017deep,Thuerey2018,zhang2018application}.
The evaluation of these networks for prediction is considerably faster than traditional numerical PDE solvers, allowing relatively accurate solutions to be predicted between one and four orders of magnitude faster \citep{guo2016convolutional,farimani2017deep}.
For this reason, CNNs are ideal for developing surrogate models, complementing expensive numerical solvers \citep{guo2016convolutional,miyanawala2017efficient}, or for real-time animations \citep{kim2019deep}.
Although the authors of earlier work \citep{guo2016convolutional,farimani2017deep,Thuerey2018} demonstrated the generalisation of their networks to domain geometries not seen during training, these unseen domains contain elementary geometrical entities included within the training data.
We go one step further by training the network with exclusively straight boundaries and showing the network is able to generalise to arbitrary fluid domains incorporating boundaries with varying radius of curvature.

To some extent, unsteady physics have been indirectly explored in the field of computer vision \citep{lee2019data,kim2019deep,sorteberg2018approximating,Fotiadis2020,wiewel2019latent}.
Here, the input to the network is a sequence of past solution fields, while the output is a sequence of predicted solution fields at future times.
When predicting unsteady phenomena, there is an additional challenge: keeping the predictions accurate for long time periods.
To address this, Kim et al. \cite{kim2019deep} and others \cite{sorteberg2018approximating,wiewel2019latent} proposed to use encoder-propagator-decoder architectures, whereas Lee and You \cite{lee2019data} and Fotiadis et al. \cite{Fotiadis2020} continued to use encoder-decoder architectures similar to those used for steady problems.
Time interpolation (or frame interpolation) has also been performed by neural networks recently, achieving excellent results in film and video production \cite{Liu2017,Niklaus2017,Chen2018,Bao2019}.
We take inspiration from these studies to time-interpolate our physics predictions.

\section{U-Net as a Simulation Engine}
\subsection{Method}
Our NN is based on the U-Net architecture \citep{UNet}, which has been extensively used for image-to-image translation tasks \citep{farimani2017deep,Thuerey2018,Fotiadis2020,isola2017image}.
In wave dynamics forecasting, the input sequence and the target share an important amount of information at different length scales, this makes the U-Net a particularly appropriate architecture for our problem.
Our U-Net receives six fields as input: the geometry field, $\Omega$, and a sequence of five consecutive height fields, $\{ h_{s}, h_{s+1}, h_{s+2}, h_{s+3}, h_{s+4} \}$. It generates as output a prediction of the subsequent height field, $\hat{h}_{s+5}$, at the next time point. Hence, each evaluation of the network corresponds to performing a single time-step, and the network is re-fed with past predictions to make further predictions. 
See Appendix \ref{sec:unet} for more details about our U-Net architecture.

\begin{figure}[h]
     \centering
     \begin{adjustbox}{minipage=\linewidth,scale=0.9}
     \centering
     \begin{subfigure}[b]{0.22\textwidth}
         \centering
         \includegraphics[width=\textwidth]{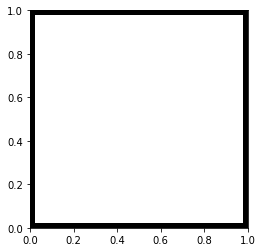}
         \caption{\textbf{Square box (tr.)}}
         \label{fig:box}
     \end{subfigure}
     \begin{subfigure}[b]{0.22\textwidth}
         \centering
         \includegraphics[width=\textwidth]{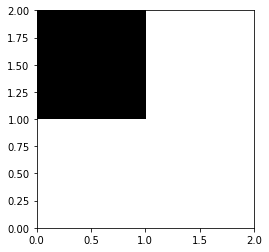}
         \caption{\textbf{Corner (tr.)}}
         \label{fig:corner}
     \end{subfigure}
     \begin{subfigure}[b]{0.22\textwidth}
         \centering
         \includegraphics[width=\textwidth]{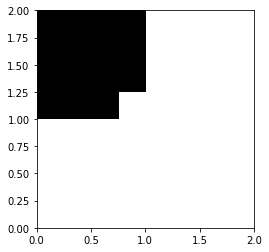}
         \caption{Double corner}
         \label{fig:steps}
     \end{subfigure}
     \begin{subfigure}[b]{0.22\textwidth}
         \centering
         \includegraphics[width=\textwidth]{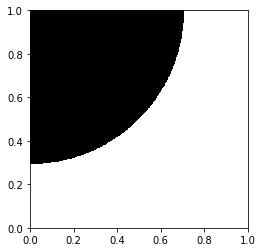}
         \caption{Convex circle}
         \label{fig:convex_circle}
     \end{subfigure}
     \begin{subfigure}[b]{0.22\textwidth}
         \centering
         \includegraphics[width=\textwidth]{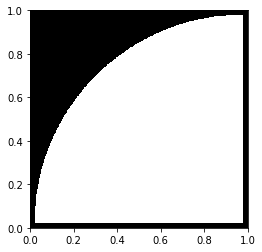}
         \caption{Concave circle}
         \label{fig:concave_circle}
     \end{subfigure}
     \begin{subfigure}[b]{0.22\textwidth}
         \centering
         \includegraphics[width=\textwidth]{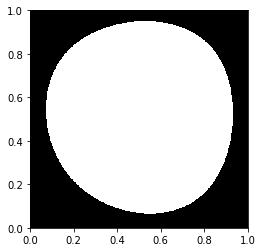}
         \caption{B-Splines}
         \label{fig:splines}
     \end{subfigure}
      \begin{subfigure}[b]{0.22\textwidth}
         \centering
         \includegraphics[width=\textwidth]{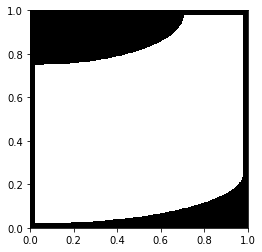}
         \caption{Ellipse arcs}
         \label{fig:ellipse}
     \end{subfigure}
    \end{adjustbox}
    \caption{Flow domains on our training (\textbf{tr.}) and testing sets. Dimensions in metres.}
    \label{fig:omega}
\end{figure}

The data sets used during training and testing were generated by solving the inviscid, two-dimensional shallow water equations with Nektar++, a high-order spectral/$hp$ element solver \citep{nektar}.
We imposed two forms of boundary conditions: solid wall boundaries, which result in wave reflection and diffraction; and open boundaries, which allow waves to exit the domain.
As initial conditions, we considered a \emph{droplet}, represented mathematically by a localised two-dimensional Gaussian, randomly placed inside the domain.
Figure \ref{fig:omega} shows the seven categories of fluid domains included in the data sets. For full details of the data sets, see Appendix \ref{sec:data}.
Our U-Net was trained against the simple closed box and open corner geometries shown in Figures \ref{fig:box} and \ref{fig:corner} for five time-steps.
The time step was set to $\Delta t = 0.12$ s and the spatial resolution was set to 128 pix/m (details in Appendix \ref{sec:unet}.1)

\subsection{Generalisation to Different Domain Geometries}
\begin{figure}[h]
\centering
\includegraphics[clip,width=0.5\columnwidth, trim={30mm 45mm 15mm 45mm}]{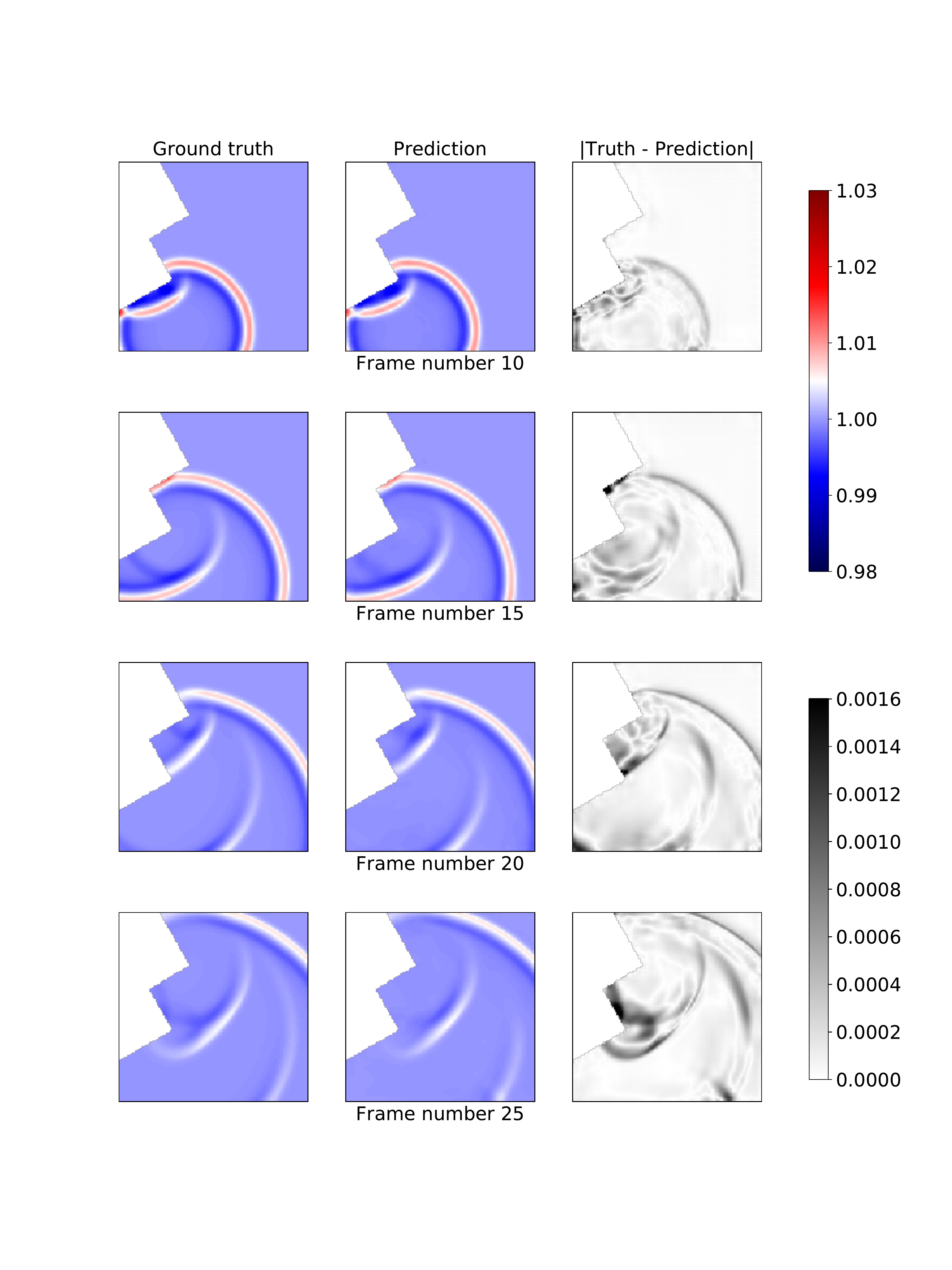}
\caption{Ground truth (left column), predictions (centre) and absolute difference (right) for a droplet as initial condition [\href{https://ibb.co/jfWqSdW}{\underline{video}}].}
\label{fig:res_steps}
\end{figure}

Our network proved to generalise to complex domain geometries not seen during training, some including curved walls, when only the closed domain and open-corner domain were used for training (Figures \ref{fig:box} and \ref{fig:corner}).
Figure \ref{fig:res_steps} shows that our network is able to accurately predict the wave speed and correctly infer the appearance of reflections at the solid boundaries.
The main discrepancy between ground truth and predictions is the height of the wavefronts diffracted by the edges of the steps, which are of lower magnitude in the predicted fields.
The reason for this may be that the network is less exposed to wave diffraction than to wave propagation and reflection during training.

The network is also able to produce good-quality predictions for configurations involving reflections on walls whose radii of curvature is not uniform (see Figure \ref{fig:intro}), although in these cases more substantial differences between targets and predictions can be seen.
The successful generalisation to this kind of domain is likely due to the localised support of the convolution kernels and the architecture.
When the convolutions are applied, the network may interpret curved walls as polygonal walls made up of lots of many straight walls of size equal to the stride size.
In that case, the higher the image resolution, the smaller the radius of curvature that could be handled by the network.\footnote{Links to animations comparing the ground truth and the U-Net predictions on the fluid domains in Figure \ref{fig:omega} can be found in \href{https://doi.org/10.6084/m9.figshare.13182623.v1}{https://doi.org/10.6084/m9.figshare.13182623.v1}}

\section{3D CNN for Time-Interpolation}
\subsection{Method}
To perform the time interpolation we opted to use a 3D CNN, since 3D convolutional kernels could capture the spatio-temporal dependencies of our problem.
We designed our 3D CNN in such a way that the time-step size of the output sequence is four times smaller than in the input sequence, i.e., three temporal frames are inserted between two original temporal frames.
Basically, our 3D CNN consists of seven 3D convolutional layers and two transposed convolutional layers (see Appendix \ref{sec:3dcnn} for details).
We trained our network with the same simulations used for training the U-Net, but this time adding noise of magnitude similar to the error of the U-Net predictions.
The network was trained against simulations with five time-points, returning as output 17 time-points per simulation.

\subsection{Results}
Figure \ref{fig:comp3dcnn} compares, at a certain time point, the ground truth and the time-interpolations obtained after feeding the U-Net predictions to our 3D CNN and a linear interpolator.
We can appreciate an acceptable similarity between the ground truth and the prediction of our (U-Net)-(3D-CNN) model, especially compared to the (U-Net)-(linear interpolator) model.
In addition, we can conclude that the discrepancies with the ground truth introduced by our 3D CNN into the U-Net predictions are almost irrelevant.
This is evidenced in Figure \ref{fig:l13dcnn}, where it can be seen that the L$_1$ error of the U-Net and the L$_1$ error of the (U-Net)-(3D-CNN) system are of the same order.

\begin{figure}[h]
\centering
\begin{subfigure}[b]{0.55\textwidth}
         \centering
         \includegraphics[clip,width=\columnwidth]{figures/comp3d.pdf}
         \caption{U-Net prediction of the wave dynamics (left), time-interpolation of such prediction performed by our network (centre) and linear interpolation (right) [\href{https://ibb.co/xf1rJSG}{\underline{video}}].}
         \label{fig:comp3dcnn}
     \end{subfigure}
     \hspace{3mm}
     \begin{subfigure}[b]{0.4\textwidth}
         \centering
         \includegraphics[width=\textwidth]{figures/l13dcnn.pdf}
         \caption{Time point vs L1 error between the original U-Net predictions, after feeding the 3D-CNN and after linear interpolating.}
         \label{fig:l13dcnn}
     \end{subfigure}
\caption{}
\label{fig:time}
\end{figure}

\section{Conclusion}
In this work, we investigated the application of fully convolutional deep neural networks for forecasting wave dynamics on fluid surfaces. In particular, we focused on a U-Net architecture with two skip connections,
and we trained the network to predict the spatio-temporal evolution of wave dynamics, including: wave propagation, wave interference, wave reflection and wave diffraction.
% %
The domains considered during training only included a closed box and a single right-angled corners. However, our U-Net was able to extrapolate to curved walls with varying radii of curvature.
The RMSE was of order 0.0001 times the characteristic length for at least 20 time steps for both the training and testing datasets.
When run on a GPU, these simulations are around $10^4$ times faster than the equivalent numerical simulation used for generating our data sets.
These findings highlight the potential for neural networks to accurately approximate the evolution of wave dynamics with computational times several orders of magnitude smaller than conventional numerical simulation.
Additionally, we proposed to perform deep-learning-based time-interpolation to reduce the time-step size of the simulations, and we used a 3D CNN to accurately increase the number of time points of the U-Net predictions by a factor of four.
This approach makes possible to modify the time-step size of the predictions without re-training our physical model.

\FloatBarrier
\bibliographystyle{unsrt}
\bibliography{bib.bib}

\appendix

\section{U-Net Architecture} \label{sec:unet}
\begin{figure}[ht]
\centering
\includegraphics[width=0.8\textwidth]{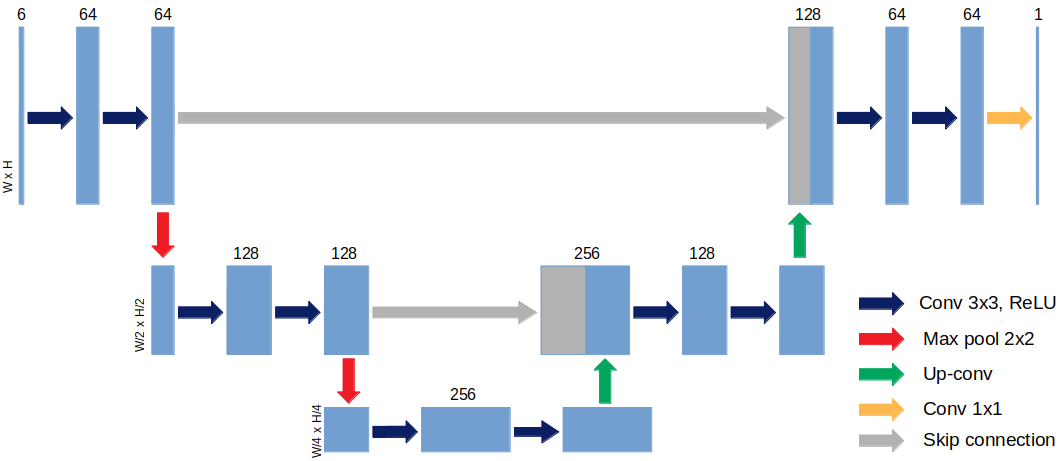}
\caption{Our U-net architecture with 1,864,577 trainable parameters. It receives six fields as input: the geometry field and a sequence of five consecutive height fields. The output is a prediction of the height field in the subsequent time step.}
\label{fig:unet}
\end{figure}

The diagram in Figure \ref{fig:unet} depicts the U-Net architecture used in the present work.
The network receives six fields as input: the geometry field $\Omega$ and a sequence of five consecutive height fields $\{ h_{s}, h_{s+1}, h_{s+2}, h_{s+3}, h_{s+4} \}$.
The output is a prediction of the subsequent height field $\hat{h}_{s+5}$.
Whereas some recent studies \citep{Fotiadis2020,Thuerey2018} have used bi-linear interpolation to perform the up-sampling, we opted to use transpose convolutions with a 2x2 kernel and stride 2. This increases the number of trainable parameters to 1,864,577, but we also noticed a significant improvement in the quality of the predictions.

\subsection{Training}
Our U-Net was trained against the simple closed box and open corner geometries shown in Figures \ref{fig:box} and \ref{fig:corner}, with a single droplet as initial condition.
The time step was set to $\Delta t = 0.12$ seconds and the spatial resolution was set to 128 pix/m.
A series of transformations were applied to perform data augmentation and normalisation (see Appendix \ref{sec:aug}).
We trained for 500 epochs with the Adam optimiser and its standard parameters \citep{kingma2014adam} and the loss function given by 
\begin{dmath} \label{eq:GradLoss}
    \mathcal{L} = (1-\lambda)\mathrm{MSE}(\hat{h}_i,h_i) + \lambda\bigg[ \mathrm{MSE}\bigg(\frac{\partial \hat{h}_i}{\partial x}, \frac{\partial h_i}{\partial x}\bigg) + \mathrm{MSE}\bigg(\frac{\partial \hat{h}_i}{\partial y},\frac{\partial  h_i}{\partial y}\bigg) \bigg]
\end{dmath} 
with $\lambda=0.05$.
The learning rate was set to $10^{-4}$ and decreased by a factor of 10 every 100 epochs.
The time origin for the input sequences is not at $t=0$, instead it is randomly selected, and five time-steps are performed before updating the network weights, whereas the loss function is evaluated after every time-step using backpropagation through time.

\section{Data Sets} \label{sec:data}
All data sets used during training and testing were generated by solving the inviscid, two-dimensional shallow water equations with Nektar++, a high-order finite element solver \cite{nektar}.
In conservative form, these equations are given by
\begin{equation}
\frac{\partial}{\partial t}
\begin{pmatrix}
h \\
hu \\
hv
\end{pmatrix}
+ \nabla \cdot
\begin{pmatrix}
hu & hv\\
hu^2 + gh^2/2 & hvu\\
huv & hv^2 + gh^2/2
\end{pmatrix}
=
\textbf{0},
\ \ \ \ (x,y)\in \mathcal{D}
\end{equation}
where $g = 9.80665 \text{ m/s}^2$ is the acceleration due to gravity and $\mathcal{D} \in \R^2$ denotes the flow domain under consideration.
Unknown variables in this system are the water depth $h(x,y,t)$ and the components of the two-dimensional velocity vector $u(x,y,t)$ and $v(x,y,t)$.

We imposed two forms of boundary conditions: solid wall boundaries, which result in wave reflection, diffraction and interference; and open boundaries, which allow the wave to exit the domain.
As initial conditions, we considered a \emph{droplet}, represented mathematically by a localized two-dimensional Gaussian superimposed on a unitary depth:
\begin{equation} \label{eq:ic1}
h_0^1 = 1 + I\exp\bigg(-C\big((x-x_c)^2+(y-y_c)^2\big)\bigg)
\end{equation}
where $I$ is set to 0.1 m and the values of $C$, $x_c$ and $y_c$ are randomly selected for each simulation.
$C$ follows a uniform distribution such that $C \in (400, 1000) \text{ m}^{-2}$.
Similarly, the centre of the Gaussian follows a two-dimensional uniform distribution such that $(x_c, y_c) \in \mathcal{D}$.

The sequences in each data set contain 100 snapshots of the height field sampled at intervals of $\Delta t = 0.03$ seconds.
%These frames are stored as 512$\times$512 arrays.
Each sequence is associated with a binary geometry field, $\Omega(x,y)$, which satisfies
\begin{equation} \label{eq:omega}
\Omega(x,y)= 
\begin{cases}
    0, & \text{if } (x,y) \in \mathcal{D} \\
    1, & \text{otherwise}
\end{cases}
\end{equation}
Therefore, $\Omega = 0$ inside the fluid domain and $\Omega = 1$ outside the fluid domain (solid boundaries).
The geometry field is an additional input to the network, which provides the required information about the boundaries. 
Figure \ref{fig:omega} shows the geometry field for the seven classes of fluid domains included in the data sets, Table \ref{table:datasets} describes these data sets.

\begin{table}[ht]
\centering
\caption{Training and testing data sets.}
\begin{tabular}{llcccc} 
 \toprule
ID & Dataset         & Purpose  & Geometry      & Initial Condition &  Sequences \\ 
 \midrule
A & Box\_Single\_Drop    & Training & Figure \ref{fig:box}    & Single Drop (eq. (\ref{eq:ic1}))       &  500       \\
B & Corner\_Single\_Drop & Training & Figure \ref{fig:corner} & Single Drop (eq. (\ref{eq:ic1}))       &  500       \\
C & Steps\_Single\_Drop  & Testing & Figure \ref{fig:steps} & Single Drop (eq. (\ref{eq:ic1}))       &  200       \\
D & Convex\_Single\_Drop & Testing & Figure \ref{fig:convex_circle} & Single Drop (eq. (\ref{eq:ic1}))       &  250       \\
E & Concave\_Single\_Drop & Testing & Figure \ref{fig:concave_circle} & Single Drop (eq. (\ref{eq:ic1}))       &  500       \\
F & Spline\_Single\_Drop & Testing & Figure \ref{fig:splines} & Single Drop (eq. (\ref{eq:ic1}))       &  200       \\
G & Ellipse\_Single\_Drop & Testing & Figure \ref{fig:ellipse} & Single Drop (eq. (\ref{eq:ic1}))       &  200       \\
\bottomrule
\label{table:datasets}
\end{tabular}
\end{table}

The data sets A-C contain only straight-sided fluid domains.
Data set A only includes wall boundary conditions, whereas B and C include open boundaries.
Data sets D-G contain curved-sided domains.
The domains in data set F were generated randomly given four random control points and using B-Splines to create a closed domain.
The domains in data set G were also generated randomly given a concave quarter of an ellipse and a convex quarter of an ellipse, whose minor axes follow a uniform distribution between 0.25 and 0.5 m.
Only the data sets A and B were used during training.
The remaining data sets are used to demonstrate the ability of the network to generalise to arbitrary fluid domain geometries and to two droplets as initial condition.

\subsection{Normalisation and data augmentation} \label{sec:aug}
To improve generalisation across a range of wave dynamics, the height fields were re-scaled according to $ \tilde{h} \leftarrow (h-\bar{h})/(\max(h)-\bar{h}) $, where $\bar{h}=1$ and $\max(h)=1.1$.
This re-scaling was reversed for visualising the network predictions.
In order to avoid over-fitting and improve the generalisation capabilities of the network, we apply two sets of transformations to the training data sets.
For the data set \textit{A} such transformations consist on random rotations of 90, 180 and 270 deg as well as horizontal and vertical flips.
For the dataset \textit{B} random rotations by multiples of 15 deg are applied and the physical size of the frames is reduced to 1 m $\times$ 1 m by cropping the original frames to domains whose center position follows an uniform distribution from 0.9 to 1.1 m in both spatial directions.
Finally, the images of all these sequences were linearly interpolated to a 128 $\times$ 128 resolution to satisfy the 128 pix/m requirement.
Regarding the testing data sets, sets \textit{C, D} are augmented in the same manner as the data set \textit{A}, since they contain open boundaries; and data sets \textit{E, F, G} are augmented like \textit{B}, as they contain only closed boundaries.

% \section{Simulation Examples} \label{sec:examples}
% Videos comparing Nektar++ simulations and simulations ran with our U-Net can be accessed on the links in Table \ref{table:gifs}.

% \begin{table}[ht]
% \centering
% \caption{Links to simulation examples}
% \begin{tabular}{llc} 
%  \toprule
% ID & Dataset         & Link \\ 
%  \midrule
% A & Box\_Single\_Drop    &   \href{https://ibb.co/pWhwrbd}{https://ibb.co/pWhwrbd}    \\
% B & Corner\_Single\_Drop &   \href{https://ibb.co/BZ0z1QC}{https://ibb.co/BZ0z1QC}   \\
% C & Steps\_Single\_Drop  & \href{https://ibb.co/jfWqSdW}{https://ibb.co/jfWqSdW} \\
% D & Convex\_Single\_Drop &  \href{https://ibb.co/PYVVp6V}{https://ibb.co/PYVVp6V}    \\
% E & Concave\_Single\_Drop &   \href{https://ibb.co/71bQMS5}{https://ibb.co/71bQMS5}    \\
% F & Spline\_Single\_Drop &    \href{https://ibb.co/Cw4m4wt}{https://ibb.co/Cw4m4wt}   \\
% G & Ellipse\_Single\_Drop &  \href{https://ibb.co/thjW5n4}{https://ibb.co/thjW5n4}     \\
% \bottomrule
% \label{table:gifs}
% \end{tabular}
% \end{table}

\section{3D-CNN Architecture} \label{sec:3dcnn}
The 3D CNN used for time-interpolating the physics predictions consists of seven convolutional layers and two transposed convolutional layers with self-normalised ELU activation functions (except the last layer).
Table \ref{table:3dcnn} summarises the hyper-parameters of these layers.

\begin{table}[ht]
\centering
\caption{3D-CNN layers.}
\begin{tabular}{cccccc} 
 \toprule
Layer & Type & Output channels & Kernel size & Stride & Padding \\ 
 \midrule
1 & 3DConv  & 32 & 3x3x3 & 1, 1, 1 & 1, 1, 1 \\
2 & 3DConv  & 32 & 3x3x3 & 1, 1, 1 & 1, 1, 1 \\
3 & 3DTConv & 32 & 2x3x3 & 2, 1, 1 & 0, 1, 1 \\
4 & 3DConv  & 64 & 2x3x3 & 1, 1, 1 & 0, 1, 1 \\
5 & 3DConv  & 64 & 3x3x3 & 1, 1, 1 & 1, 1, 1 \\
6 & 3DTConv & 64 & 2x3x3 & 2, 1, 1 & 0, 1, 1 \\
7 & 3DConv  & 64 & 2x3x3 & 1, 1, 1 & 0, 1, 1 \\
8 & 3DConv  & 32 & 3x3x3 & 1, 1, 1 & 1, 1, 1 \\
9 & 3DConv  &  1 & 1x1x1 & 1, 1, 1 & 0, 0, 0 \\
\bottomrule
\label{table:3dcnn}
\end{tabular}
\end{table}

\end{document}